\documentclass[10pt,twocolumn,letterpaper]{article}

\usepackage{cvpr}              
\usepackage{multirow}

\renewcommand{\paragraph}[1]{\vspace{.1em}\noindent\textbf{#1.}}

\definecolor{cvprblue}{rgb}{0.21,0.49,0.74}
\usepackage[pagebackref,breaklinks,colorlinks,allcolors=cvprblue]{hyperref}

\newcommand{\nickname}{LetCamsGo}
\newcommand{\best}[1]{\colorbox{red!25}{#1}}
\newcommand{\second}[1]{\colorbox{yellow!35}{#1}}
\newcommand{\none}[1]{\colorbox{white}{#1}}
\usepackage{gensymb} 
\usepackage{cuted}
\usepackage{capt-of}
\usepackage{booktabs}   
\usepackage{makecell}   
\usepackage{pifont}     
\usepackage{xcolor}     
\newcommand{\cmark}{\textcolor{green!50!black}{\ding{51}}} 
\newcommand{\xmark}{\textcolor{red}{\ding{55}}}            

\definecolor{lightblue}{rgb}{0.0, 0.75, 0.96}
\definecolor{nicegreen}{rgb}{0.13, 0.7, 0.13}
\definecolor{darkgreen}{rgb}{ .0, .5, .0}

\def\paperID{20} 
\def\confName{4DV}
\def\confYear{2026}

\title{4D Reconstruction from Sparse Dynamic Cameras}

\author{
Kazuki Ozeki$^{1}$ \quad
Shun Kenney$^{1}$ \quad
Yuto Shibata$^{1}$ \quad
Eisuke Takeuchi$^{1}$ \quad
Takuya Narihira$^{2}$ \\
Kazumi Fukuda$^{2}$ \quad
Ryosuke Sawata$^{2}$ \quad
Yuki Mitsufuji$^{2, 3}$ \quad
Yoshimitsu Aoki$^{1}$ \\
$^{1}$Keio University \quad
$^{2}$Sony AI \quad
$^{3}$Sony Group Corporation
}

\begin{document}
\maketitle
\begin{strip}
  \centering
  \includegraphics[width=\linewidth]{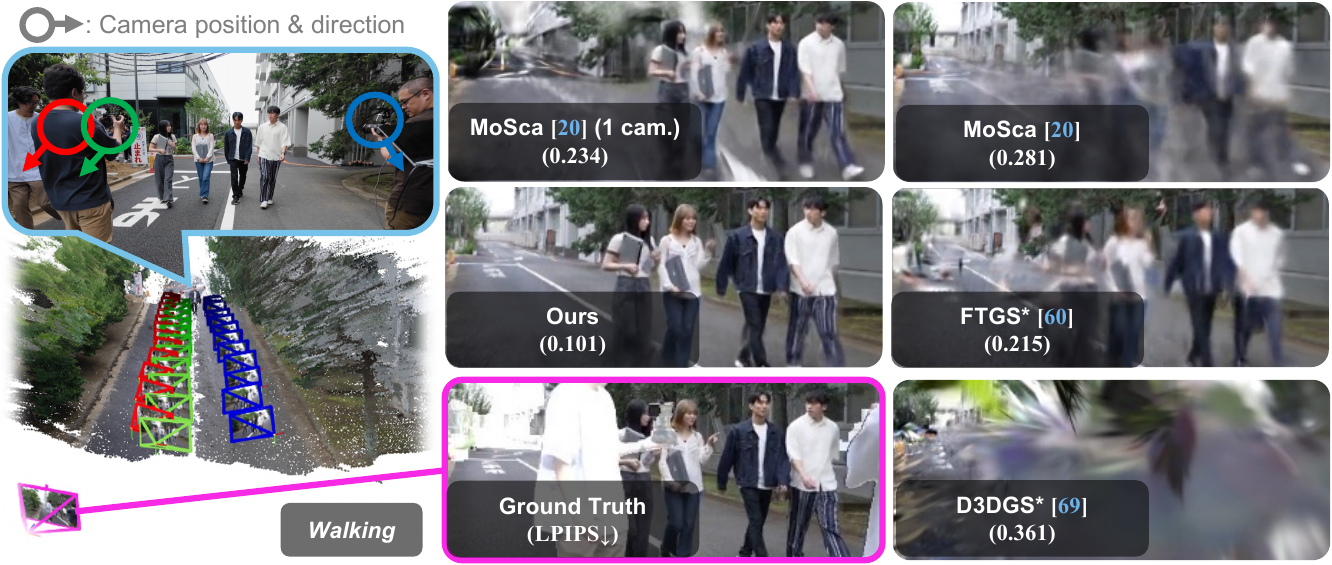}
  \captionof{figure}
  {\textbf{Overview.} 
  We introduce the task of 4D reconstruction from \emph{sparse dynamic cameras}, 
  a versatile and depth ambiguity-free configuration of multiple, independently moving cameras prevalent in real-world video production.
  Our framework resolves the fundamental spatiotemporal inconsistencies that typically undermine naive adaptations of methods designed for monocular or dense-fixed camera setups~\cite{MoSca,FreeTimeGS,D3DGS}.
  Furthermore, we introduce \textit{\nickname{}}, a new benchmark to promote 4D reconstruction research in this camera setup. }
  \label{fig:teaser}
\end{strip}

\begin{abstract}
  Although dynamic 3D (i.e., 4D) reconstruction from a monocular dynamic camera has recently advanced, it remains fundamentally limited by depth ambiguity. 
  In this paper, we focus on an alternative practical way, i.e., sparse dynamic camera setup, where a handful of independently moving cameras capture the same subjects.
  While keeping capture costs low, this setup introduces multi-view constraints and remains practical for real-world video production such as sports, concerts, and TV shows. 
  Despite its potential, our experiments show that naive extensions of existing monocular or dense-fixed camera-based methods are insufficient since they fail to resolve the complex spatiotemporal inconsistencies across views and time.
  To fill this gap, we propose a simple yet effective 3D track initialization method designed to ensure spatiotemporal consistency by integrating inter-camera feature matching with intra-camera point tracking.
  Additionally, we incorporate a noise-robust depth-ordering regularization loss and a spatiotemporally diverse batch sampling strategy to enhance optimization stability and cross-view generalization.
  Furthermore, to address the lack of standardized benchmarks for this task, we introduce \nickname{}, a new real-world video dataset with 5 sequences across 4 diverse environments, recorded by three independently moving cameras and one fixed camera. 
  Comprehensive benchmarking on \nickname{} demonstrated that our proposed framework improves 4D reconstruction quality in dynamic regions compared with baselines, paving the way for a low-cost 4D reconstruction paradigm in the wild.
\end{abstract}
    
\section{Introduction}
\label{sec:intro}

Dynamic 3D (i.e., 4D) reconstruction from a monocular dynamic RGB video has become increasingly viable with the progress of 3D representations~\cite{NeRF, 3DGS} and 2D foundation models~\cite{depthanything3, CoTracker3, SAM2}. Compared with the traditional dense fixed camera setup, this setup greatly simplifies data capture, as it only requires a single camera to move around the scene. However, it inherently suffers from depth ambiguity due to the absence of multi-view constraints, even when state-of-the-art methods~\cite{MoSca, SOM, HiMoR} are applied.

To achieve high-fidelity reconstruction while minimizing deployment complexity and capture costs, we focus on the \textit{sparse dynamic camera} setup, where a handful of independently moving cameras capture the same subjects. It can resolve depth ambiguity with multi-view constraints while remaining easier to set up than dense fixed cameras (often exceeding 20 units). Additionally, this setup is practical and commonly seen in real-world video production, including live concerts, TV show productions, and casual recordings captured by multiple smartphones. By reconstructing 4D scenes from these recordings, creators can generate free-viewpoint videos that offer more immersive viewing experiences and enable efficient post-production editing.

Despite these potentials, 4D reconstruction under the sparse dynamic camera setup has remained underexplored. We hypothesize that this is mainly due to two reasons: (i) Spatiotemporal inconsistencies in reconstruction. As demonstrated by our experiments, naive extensions of existing methods tailored for monocular dynamic cameras, such as MoSca~\cite{MoSca}, significantly degrade reconstruction quality.
This degradation stems from spatiotemporal inconsistencies in estimated geometry and correspondences, which persist even when incorporating multi-view depth estimators~\cite{depthanything3} and 3D point trackers~\cite{rajic2025mvtracker}.
(ii) The absence of standardized benchmarks. 
While the setup offers immense flexibility, its unconstrained nature—characterized by arbitrary camera orientations and complex baseline dynamics—makes it difficult to establish a rigorous evaluation protocol.

Motivated by the aforementioned discussion, this paper presents the effective 4D reconstruction framework for the sparse dynamic camera setup. Specifically, we propose a simple yet effective 3D track initialization method that combines inter-camera feature matching with intra-camera point tracking by leveraging pre-trained vision foundation models~\cite{GIM,MASt3R,CoTracker3}. 
This integration establishes dense spatiotemporal 2D correspondences across dynamic regions. By subsequently applying epipolar filtering and per-frame triangulation, our method derives 3D tracks that maintain strict multi-view consistency, providing a reliable motion-scaffold initialization.
Furthermore, to stabilize optimization in this unconstrained setup, we incorporate a depth order regularization loss and introduce a spatio-temporal batch sampling strategy designed to capture cross-view and cross-temporal variations.

In addition, we introduce \emph{\nickname}{}, a new real-world video dataset capturing dynamic scenes with three independently moving cameras and one fixed camera for this novel task (see~\cref{fig:teaser}). 
\nickname{} comprises 5 sequences across 4 diverse indoor and outdoor environments, featuring multiple subjects performing everyday activities with non-human object interactions, complex non-rigid motions, and frequent occlusions to provide realistic and challenging 4D reconstruction scenarios. 
To facilitate a rigorous and systematic evaluation, we establish a comprehensive taxonomy for \nickname{} based on four critical axes: the spatial extent of subject motion, camera orientation, motion direction, and the temporal dynamics of camera baselines. 
This structured classification ensures that our benchmark provides a granular look at how different capture conditions impact 4D reconstruction fidelity.

To analyze the unique features and challenges of the sparse dynamic camera setup, we compare our method not only with a naive extension of MoSca but also with various baselines including FreetimeGS~\cite{FreeTimeGS}, a state-of-the-art method for dense fixed cameras. Through benchmarking on \nickname{}, we demonstrate that our approach significantly outperforms the naive MoSca-based extension in most scenes and achieves superior temporal consistency in dynamic regions compared with FreetimeGS. Comprehensive experiments, including ablation studies, reveal the potential of the sparse dynamic camera setup, and we believe that \nickname{} and our method will facilitate future research in this direction.

In summary, our contributions are as follows: 
(1) We present a unified framework for the sparse dynamic camera setup. It integrates a spatiotemporally consistent 3D track initialization, a robust depth regularization loss, and a spatiotemporal batch sampling strategy.
(2) We introduce \nickname{}, a real-world dataset for 4D reconstruction in the sparse dynamic camera setup. 
(3) We provide comprehensive benchmarking and analysis on \nickname{}, showing the effectiveness of our method and highlighting the challenges and potential of the sparse dynamic camera setup.

\section{Related Work}
\label{sec:related_work}


\subsection{4D Reconstruction}
\label{subsec:related_work_4d_recon}

Before extending to 4D, there are various 3D representation methods; classic and widely used representations include point clouds, meshes, and voxels. 
With the recent rise of Neural Radiance Fields (NeRF)~\cite{NeRF} and 3D Gaussian Splatting (3DGS)~\cite{3DGS}, end-to-end differentiable optimization of spatial representations has become possible. In particular, 3DGS has attracted attention as a tool that achieves both fast optimization and rendering while maintaining high fidelity. 
More recently, applying 3DGS to 4D scenes containing dynamic objects has been a hot research topic, and thus many methods have been proposed. 
The temporal representation of Gaussians for reconstructing 4D scenes includes: methods~\cite{D3DGS, D2GV, EvolvingGS, SplatVoxel, GIFStream, InstantGS, QUEEN, PerGauEBD, Dynamic3DG, 3DGeoawareDefGS, 4DGS, SplineGSsmooth, SOM, SplatteraVideo, SplatFields, 4DGC, ODHSR} that model temporal deformation from Gaussians defined in a canonical space, methods~\cite{STG, FreeTimeGS, HiMoR, RealtimePDCR, 4DGaussian} that place Gaussians in a shared spacetime and model time-varying opacity to capture scene dynamics, and hybrid approaches~\cite{MoSca, GauSTAR, 4DTAM, 4D-Fly, SCGS} that combine these ideas.

Meanwhile, feed-forward approaches~\cite{MoVieS, l4GM, InstantGS, BTimer, GPSGaussian, Geo4D} have emerged that directly output Gaussian parameters using learned deep learning models. While these methods enable instant reconstruction, many still fall short in accuracy compared to meticulous optimization. Moreover, most existing work targets scenes captured by monocular video, and there are relatively few attempts to reconstruct dynamically captured multi-camera scenes. Concurrent work~\cite{HMRoSP} also tackle with the sparse dynamic camera setup, yet it only focuses reconstruct human mesh.\\


\subsection{Datasets for 4D Reconstruction}
\label{subsec:related_work_4d_data}
Existing real-world video datasets for 4D reconstruction can be broadly categorized by camera setup, namely the number of cameras and their motion patterns. Dynamic-camera datasets~\cite{DAVIS,3DPW,DyCheck,WildAvatar,DynPose,MoDGS} are generally limited to a single moving camera. In contrast, multi-camera datasets~\cite{NVIDIA,Neural3DV,StreamRF,MotionX,Technicolor,Immersive,PanopticStudio,ENeRF,FreeMan,SelfCap,MVTrack,GIFStream,ImViD} mostly use fixed cameras. Some datasets~\cite{ImViD, nuScenes, Waymo, Ego4D, ReCamMaster, Ego4D, ReCamMaster} include multiple dynamic cameras, but they differ from our setting because the cameras are rig-mounted, capture different subjects, or are simulated. A detailed comparison is provided in the supplementary material. The closest camera setup to ours is used in~\cite{4DRecon}, which features five independently moving cameras; however, each camera is quasi-static, making the setup closer to sparse fixed-camera capture. By contrast, our dataset contains widely moving cameras that cover larger scenes with fewer cameras (see~\cref{fig:pose}). This setting provides more immersive experiences and is more practical for real-world applications, while also introducing greater spatiotemporal inconsistencies due to large viewpoint changes.

\section{Method}
\label{sec:method}
Our goal is to recover a high-fidelity 4D representation from time-synchronized RGB videos captured by multiple independently moving cameras. Each camera \#$c$ $(= 1,2,\cdots,C;~C\mbox{ is the number of cameras})$ has a known intrinsic matrix $\mathbf{K}_{c}$ and time-varying extrinsic parameters $\{\mathbf{R}_{c,t},\mathbf{t}_{c,t}\}_{t=1}^{T}$. 
To apply our frameworks to the sparse dynamic camera setup, we employ MoSca~\cite{MoSca} as the backbone model due to its strong reconstruction quality in the monocular dynamic camera setup.
On the basis of MosCa, we
focus on improving its motion-scaffold initialization for the sparse dynamic camera setup.

We first review MoSca in~\cref{subsec:preliminary} and \cref{subsec:naive} discusses a naive multi-view extension of MoSca and its limitations. \cref{subsec:3d_track} then presents our multi-view consistent 3D track initialization, and \cref{subsec:optimization} introduces additional effective optimization strategies.

\begin{figure}
  \centering
  \begin{subfigure}{0.49\linewidth}
    \includegraphics[width=\linewidth]{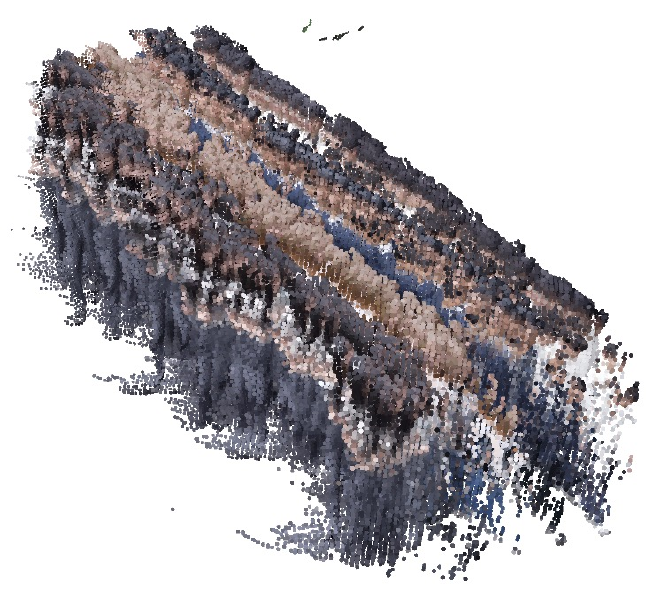}
    \caption{Naive Approach}
    \label{subfig:naive_mvtracker}
  \end{subfigure}
  \hfill
  \begin{subfigure}{0.49\linewidth}
    \includegraphics[width=\linewidth]{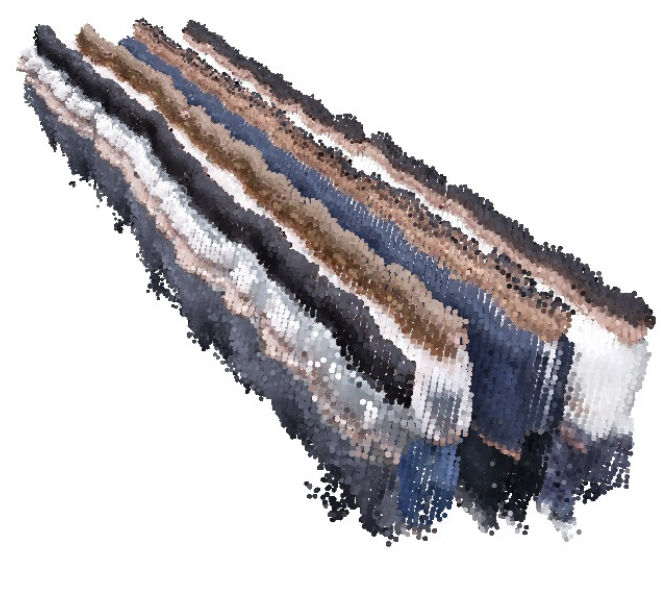}
    \caption{Ours}
    \label{subfig:naive_ours}
  \end{subfigure}
  \caption{\textbf{Initial dynamic points.} Pose-conditioned metric multi-view depth estimation~\cite{depthanything3} and multi-view 3D point tracking~\cite{rajic2025mvtracker} produce noisy and inconsistent dynamic points across times and views (a). In contrast, our multi-view consistent 3D track initialization produces more accurate and consistent dynamic points (b).}
  \label{fig:naive}
\end{figure}

\begin{figure*}[t]
  \includegraphics[width=\linewidth]{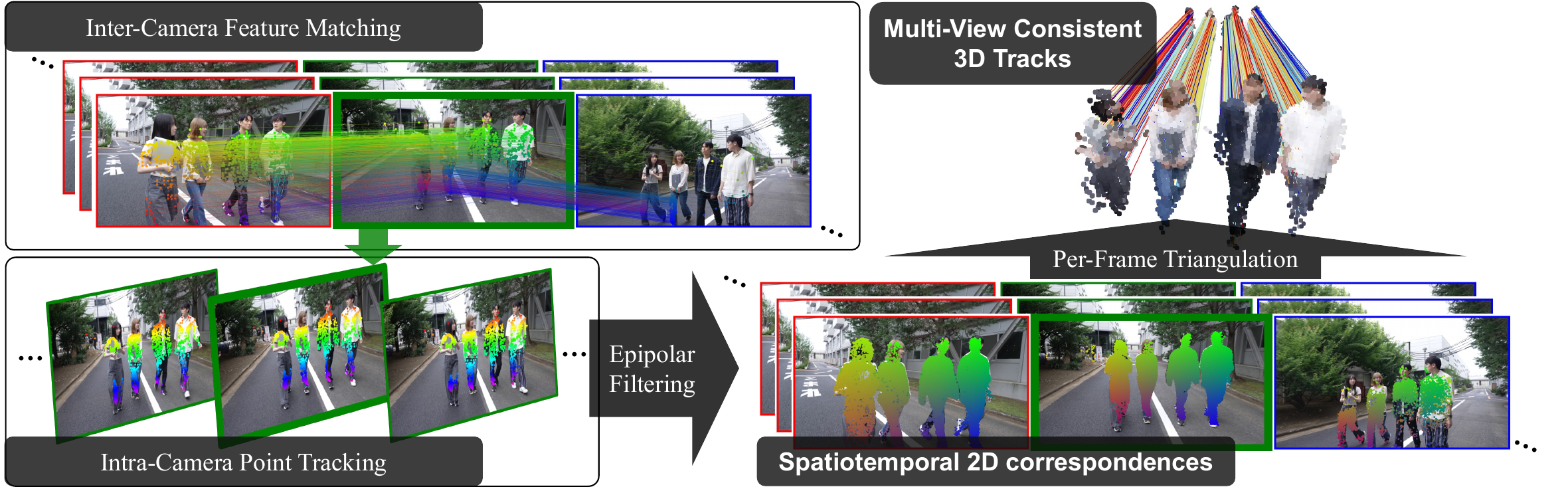}
  \caption{\textbf{Proposed multi-view consistent 3D track initialization.} We combine inter-camera feature matching and intra-camera point tracking, followed by epipolar filtering and triangulation to produce spatiotemporally consistent 3D tracks for motion-scaffold initialization.}
  \label{fig:method}
\end{figure*}

\subsection{Preliminary}
\label{subsec:preliminary}
We build on MoSca~\cite{MoSca}, which augments 3DGS~\cite{3DGS} with sparse motion-scaffold nodes. 
A scene is represented by Gaussians $\{\mathcal{G}_{i}\}$ with mean $\boldsymbol{\mu}_{i}$, covariance $\boldsymbol{\Sigma}_{i}=\mathbf{R}_{i}\mathbf{S}_{i}\mathbf{S}_{i}^{\top}\mathbf{R}_{i}^{\top}$ (where $\mathbf{S}_{i}$ is a diagonal scale matrix and $\mathbf{R}_{i}$ a unit-quaternion rotation), opacity $\alpha_{i}$, and SH color $\mathbf{h}_{i}$. After projecting to 2D and depth sorting, pixel color is composited by
\begin{equation}
  \mathbf{I}(u,v) = \sum_{i=1}^{N}\left( \alpha_{i}\, \mathbf{c}_{i} \prod_{j=1}^{i-1} (1-\alpha_{j}) \right). \label{eq:alpha_compositing}
\end{equation}
Scaffold nodes $\mathcal{V}=\{v^{(m)}\}_{m=1}^{N_m}$ carry time-varying transforms $\mathbf{Q}^{(m)}_{t}\in SE(3)$, connected in a KNN graph with motions blended by dual-quaternion blending (DQB). For query point $\mathbf{x}$ and times $(t_{s},t_{d})$, the warp is
\begin{equation}
  \mathcal{W}(\mathbf{x},t_{s}\!\rightarrow\!t_{d})=\operatorname{DQB}\!\left(\left\{\Delta \mathbf{Q}^{(m)}_{t_s\rightarrow t_d},\,w_{m}(\mathbf{x})\right\}_{m\in\mathcal{N}_K(\mathbf{x})}\right),
\end{equation}
where $\Delta \mathbf{Q}^{(m)}_{t_s\rightarrow t_d}=\mathbf{Q}^{(m)}_{t_d}(\mathbf{Q}^{(m)}_{t_s})^{-1}$ and $w_{m}(\mathbf{x})$ is a distance-based weight. Canonical dynamic Gaussians $\mathbf{g}_{n}=(\mathbf{x}_{n},\mathbf{R}_{n},\mathbf{s}_{n},o_{n},\mathbf{c}_{n},\tau_{n})$ are warped by $\mathcal{W}(\cdot,\tau_{n}\!\rightarrow\!t)$ and rendered by splatting. Parameters are optimized using a photometric loss, depth and track regularization terms, and geometric losses such as an as-rigid-as-possible scaffold loss, together with densification and pruning.

\subsection{Naive Approach}
\label{subsec:naive}
Since monocular 4D reconstruction is highly ill-posed, MoSca~\cite{MoSca} relies on 2D foundation models for initialization and supervision. In particular, the motion scaffold is initialized leveraging monocular depth estimators~\cite{depth_anything_v2,hu2024-DepthCrafter,piccinelli2024unidepth} and point trackers~\cite{CoTracker3,SpatialTracker}. A direct multi-camera extension is to run this monocular initialization independently on each camera stream and then jointly optimize all streams. In practice, this naive strategy introduces severe cross-view inconsistencies that degrade dynamic-object reconstruction quality. We observe this issue even when replacing the initialization with state-of-the-art pose-conditioned metric multi-view depth estimation~\cite{depthanything3} and multi-view 3D point tracking~\cite{rajic2025mvtracker}, likely due to the domain gap between their training data and our sparse dynamic camera setting (see~\cref{fig:naive}).

\begin{figure*}
  \includegraphics[width=\linewidth]{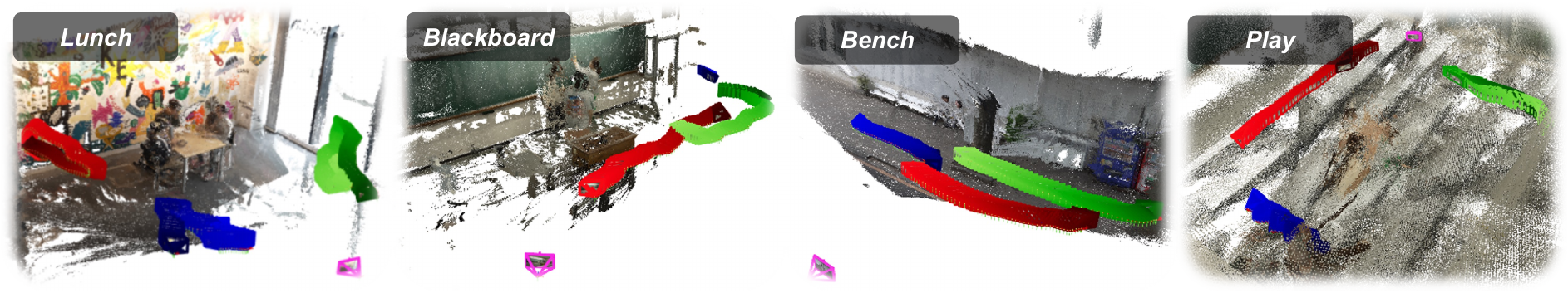}
  \caption{\textbf{Visualization of estimated camera trajectories.} The estimated trajectories of the three dynamic cameras are shown in red, green, and blue, and the fixed evaluation camera is shown in pink. For visual clarity, the frame rate is reduced to 6 FPS.}
  \label{fig:pose}
\end{figure*}

\subsection{Multi-View Consistent 3D Track Initialization}
\label{subsec:3d_track}
To address the above failure mode, we propose a multi-view consistent 3D track initialization that provides stronger priors for the motion scaffold. Our key idea is to combine \textbf{inter-camera feature matching} and \textbf{intra-camera point tracking} to propagate cross-view correspondences over time. 
Figure~\ref{fig:method} shows the overall pipeline. 

We first perform inter-camera feature matching frame by frame. Specifically, for each frame $t$, we extract 2D correspondences $(\mathcal{M}_{t}^{c},\mathcal{M}_{t}^{c'})$ between camera pairs $(c,c')$, where $N_t$ is the number of matches at frame $t$. To maximize the number of reliable correspondences, we merge outputs from two pairwise dense matchers with different architectures and training data, GIM-DKM~\cite{GIM} and MASt3R~\cite{MASt3R}. Repeating this process over all frames yields temporally indexed cross-view correspondences $\{(\mathcal{M}_{t}^{c},\mathcal{M}_{t}^{c'})\}_{t=0}^{T}$. These matches are spatially consistent within each frame, but they do not provide temporal linkage across frames, which is required for motion-scaffold initialization.
 
To establish this temporal linkage, we then perform intra-camera point tracking. For each camera $c$ and query frame $t$, we use the matched points as seeds, track them forward and backward, and merge both directions into a single trajectory over the sequence. We use CoTracker3-Online~\cite{CoTracker3}, a 2D point tracker robust to non-rigid motion, dynamic camera motion, and long-range tracking. This process produces a set of 2D tracks $\mathcal{T}_{c,t}$ for queries from frame $t$. Aggregating tracks from all query frames yields a semi-dense 2D track set $\mathcal{T}_{c}$ for each camera $c$, where each track stores 2D locations and visibility across all frames.

Given these 2D spatiotemporal correspondences, we obtain spatiotemporal 3D tracks using two additional steps, unlike the naive approach. First, we apply epipolar-geometry-based filtering to remove geometrically inconsistent cross-view correspondences. Since per-camera tracking can be noisy, mismatched cross-view pairs for the same 3D point can hinder convergence of the track regularization loss~\cite{MoSca}. For each tracked point and each time step, we evaluate whether its cross-view correspondence satisfies the epipolar constraint, and set its visibility to zero ($\nu_{i,c,t}=0$) when the constraint is violated. We use the Sampson error as the epipolar consistency metric. Given a fundamental matrix $\mathbf{F}$, we keep pairs satisfying
\begin{equation}
  d_{\mathrm{S}}(\mathbf{u},\mathbf{u}')= \frac{(\mathbf{u}'^{\top}\mathbf{F}\mathbf{u})^{2}}{(\mathbf{F}\mathbf{u})_{1}^{2}+(\mathbf{F}\mathbf{u})_{2}^{2}+(\mathbf{F}^{\top}\mathbf{u}')_{1}^{2}+(\mathbf{F}^{\top}\mathbf{u}')_{2}^{2}}<\tau_{\mathrm{epi}}.
\end{equation}

Second, we perform frame-by-frame triangulation. For each trajectory $i$ at time $t$, we estimate its 3D position by triangulating all visible observations at that frame:
\begin{equation}
  \mathbf{X}_{i,t}=\arg\min_{\mathbf{X}}\sum_{c\in\mathcal{C}_t(i)}\left\|\pi\!\left(\mathbf{P}_{c,t}\mathbf{X}\right)-\mathbf{u}_{i,c,t}\right\|_{2}^{2},
\end{equation}
where $\mathcal{C}_{t}(i)$ denotes the set of cameras observing trajectory $i$ at time $t$, $\pi(\cdot)$ is perspective projection, and $\mathbf{P}_{c,t}=\mathbf{K}_{c}\left[\mathbf{R}_{c,t}\mid\mathbf{t}_{c,t}\right]$ is the projection matrix. In contrast to back-projecting noisy depth estimates, this step directly produces multi-view consistent 3D points at each frame. Repeating it over all frames yields spatiotemporal 3D tracks, which are used to initialize the motion scaffold.

\subsection{Optimization}
\label{subsec:optimization}
\paragraph{Depth Regularization} Vanilla MoSca~\cite{MoSca} applies depth regularization by minimizing the L2 distance between the normalized estimated and rendered depth maps. However, this approach is suboptimal when the depth estimates are noisy, which is often the case in the sparse dynamic camera setup, even when using pose-conditioned metric multi-view depth estimation~\cite{depthanything3}. Instead, following~\cite{MoDGS, GC4DGS}, we regularize depth by correcting the ordering of depth values based on the estimated depth map. This strategy is more robust to depth estimation noise and yields better reconstruction quality in our setting.

\paragraph{Batch Sampling} In 3DGS-based 4D representation, batch-based optimization is often effective for improving convergence and generalization~\cite{4DGS,SOM,MonoFusion}. We find that optimization generalizes best when each batch consists of samples from \emph{different cameras} and \emph{different frames}. Compared to same-camera or temporally local batching, this strategy reduces short-range correlation in the supervision signal and introduces stronger cross-view and cross-temporal variation per update. Empirically, this approach achieves the highest reconstruction quality and best generalization in our setting.

\section{\nickname{} Dataset}
\label{sec:dataset}
To the best of our knowledge, there is no publicly available real-world video dataset under the sparse dynamic camera setup. Thus, we introduce \emph{\nickname{}}, comprising 5 sequences of realistic everyday activities across 4 diverse environments (classroom, lounge, bench, and road), each recorded with three independently moving cameras and one fixed evaluation camera. All cameras record Full HD videos at 60~FPS, time-synchronized using UltraSync devices~\cite{UltraSync}. Camera poses are estimated using COLMAP~\cite{COLMAP}, and dynamic object masks are generated semi-automatically using~\cite{SAM2,ren2024grounded}. 
Obtained camera poses for each scene are visualized in~\cref{fig:pose}.
To enable comprehensive performance analysis, we systematically categorize the 5 scenes (detailed in Table~\ref{tab:dataset_characteristics}) along four dimensions: (1) spatial extent of subject motion, (2) camera orientation, (3) camera motion, and (4) temporal dynamics of camera baselines. Further details on data acquisition, characteristics, and annotation procedures are provided in the supplementary material.

\section{Experimental Settings}
\label{sec:settings}

\begin{table}[t]
  \centering
  \caption{\textbf{Characteristics of each scene on \nickname{}.} }
  \label{tab:dataset_characteristics}
  \resizebox{\linewidth}{!}{%
  \begin{tabular}{lcccc}
    \toprule
    Scene & \makecell{Subject\\Motion} & \makecell{Camera\\Orientation} & \makecell{Camera\\Motion} & \makecell{Baseline\\Dynamics} \\
    \midrule
    Lunch      & Local                   & Forward-Facing              & Sideways               & High                         \\
    Blackboard & Local                   & Forward-Facing              & Sideways               & Low                          \\
    Play       & Local                   & Outside-In                  & Sideways               & High                         \\
    Bench      & Global                  & Forward-Facing              & Sideways               & Low                          \\
    Walking    & Global                  & Forward-Facing              & Backward               & Low                          \\
    \bottomrule
  \end{tabular}%
  }
\end{table}

\begin{table*}
  [t]
  \centering
  \caption{\textbf{Benchmarking on \nickname{}}. Each entry reports PSNR / SSIM / LPIPS. Higher is better for PSNR and SSIM, while lower is better for LPIPS. For each scene, the best and second-best values for each metric are highlighted in red and yellow, respectively.}
  \label{tab:quantitative} \resizebox{\textwidth}{!}{%
  \begin{tabular}{l*{6}{c}}
    \toprule      & \multicolumn{2}{c}{Lunch}                    & \multicolumn{2}{c}{Blackboard}               & \multicolumn{2}{c}{Play}                   \\
                  & Full                                         & Dynamic                                      & Full                                       & Dynamic                                      & Full                                    & Dynamic                                        \\
    \midrule 
    D3DGS*~\cite{D3DGS}        & \none{19.44} / \none{0.826} / \second{0.133}      & \second{20.75} / \second{0.994} / \second{0.115} & \none{18.16} / \none{0.805} / \none{0.306}                      & \none{16.13} / \second{0.973} / \none{0.286}                        & \none{14.70} / \none{0.685} / \none{0.347}                   & \none{15.45} / \none{0.979} / \second{0.267}                          \\
    FTGS~\cite{FreeTimeGS} & \none{17.20} / \none{0.790} / \none{0.187}                        & \none{18.99} / \none{0.992} / \none{0.171}                        & \none{18.74} / \none{0.819} / \none{0.298}             & \second{16.62} / \best{0.976} / \none{0.285}        & \none{16.09} / \none{0.725} / \none{0.312} & \none{15.22} / \none{0.979} / \none{0.335} \\
    FTGS*~\cite{FreeTimeGS}         & \none{20.12} / \best{0.833} / \best{0.131} & \none{19.78} / \none{0.993} / \none{0.154}      & \none{18.60} / \none{0.812} / \second{0.290}             & \best{16.83} / \best{0.976} / \second{0.277}        & \none{15.90} / \none{0.723} / \best{0.278}     & \none{14.87} / \none{0.979} / \none{0.362}          \\
    MoSca~\cite{MoSca,CoTracker3}        & \second{20.40} / \none{0.811} / \none{0.201}                        & \none{18.32} / \none{0.992} / \none{0.216}                        & \second{18.86} / \second{0.843} / \none{0.293} & \none{16.05} / \none{0.971} / \none{0.356}                        & \none{16.08} / \second{0.749} / \none{0.328}                   & \none{15.43} / \best{0.981} / \none{0.329}                          \\
    MoSca-M~\cite{MoSca,rajic2025mvtracker}        & \none{20.20} / \none{0.808} / \none{0.204}                        & \none{17.83} / \none{0.991} / \none{0.214}                        & \none{18.76} / \none{0.840} / \none{0.301} & \none{15.87} / \none{0.971} / \none{0.375}                        & \second{16.68} / \best{0.750} / \none{0.323}                   & \second{16.21} / \best{0.981} / \none{0.347}                          \\
    Ours          & \best{21.07} / \second{0.829} / \best{0.131}                 & \best{22.07} / \best{0.995} / \best{0.091}   & \best{19.23} / \best{0.845} / \best{0.277}             & \best{16.83} / \second{0.973} / \best{0.259} & \best{16.85} / \none{0.744} / \second{0.285}   & \best{16.86} / \second{0.980} / \best{0.237}          \\
    \bottomrule
  \end{tabular}
  } \resizebox{\textwidth}{!}{%
  \begin{tabular}{l*{6}{c}}
    \toprule       & \multicolumn{2}{c}{Walking}                   & \multicolumn{2}{c}{Bench}                         & \multicolumn{2}{c}{Average}                 \\
                  & Full                                           & Dynamic                                          & Full                                       & Dynamic                                 & Full                                         & Dynamic                                          \\
    \midrule
    D3DGS*~\cite{D3DGS}        & \none{12.57} / \none{0.659} / \none{0.642}                          & \none{11.07} / \none{0.989} / \none{0.361}                            & \none{16.71} / \none{0.675} / \none{0.506}    & \none{14.17} / \none{0.991} / \none{0.224}            & \none{16.32} / \none{0.730} / \none{0.387}                        & \none{15.51} / \none{0.985} / \none{0.251}                            \\
    FTGS~\cite{FreeTimeGS} & \none{14.40} / \none{0.675} / \none{0.555}                          & \none{13.71} / \none{0.991} / \none{0.278}                            & \none{18.14} / \none{0.693} / \none{0.460}                      & \best{17.14} / \best{0.994} / \second{0.168}                   & \none{16.91} / \none{0.740} / \none{0.362}                        & \none{16.34} / \second{0.986} / \none{0.247}                   \\
    FTGS*~\cite{FreeTimeGS}         & \best{18.66} / \second{0.758} / \second{0.372} & \second{16.65} / \second{0.993} / \second{0.215} & \none{16.76} / \none{0.709} / \none{0.398}                      & \second{17.12} / \second{0.993} / \none{0.170} & \none{18.01} / \none{0.767} / \second{0.294}                 & \second{17.05} / \best{0.987} / \second{0.236} \\
    MoSca~\cite{MoSca,CoTracker3}        & \none{17.23} / \none{0.751} / \none{0.481}                          & \none{12.87} / \none{0.990} / \none{0.281}                            & \none{18.82} / \second{0.733} / \second{0.394} & \none{15.33} / \none{0.991} / \none{0.195}     & \none{18.28} / \second{0.777} / \none{0.339}      & \none{15.60} / \none{0.985} / \none{0.274}                            \\
    MoSca-M~\cite{MoSca,rajic2025mvtracker}        & \none{17.05} / \none{0.743} / \none{0.498}                          & \none{12.09} / \none{0.990} / \none{0.330}                            & \second{19.26} / \best{0.735} / \none{0.401} & \none{15.80} / \none{0.991} / \none{0.201}     & \second{18.39} / \none{0.775} / \none{0.345}      & \none{15.56} / \none{0.985} / \none{0.293}                            \\
    Ours          & \second{18.45} / \best{0.766} / \best{0.360}   & \best{17.71} / \best{0.995} / \best{0.101}       & \best{19.69} / \second{0.733} / \best{0.353}             & \none{16.80} / \none{0.992} / \best{0.147} & \best{19.06} / \best{0.783} / \best{0.281} & \best{18.05} / \best{0.987} / \best{0.167}       \\
    \bottomrule
  \end{tabular}
  }
\end{table*}

\subsection{Implementation Details}
\label{subsec:implementation}
We set the Sampson error threshold $\tau_{\mathrm{epi}}$ to 0.1 px. For depth regularization, we use depth maps estimated by Depth Anything 3~\cite{depthanything3}. For simplicity, we use the dynamic mask from~\cref{sec:dataset}. Static points are initialized from off-the-shelf multi-view stereo depth maps~\cite{COLMAP} for all scenes except \textit{Play}, where they are initialized from depth maps estimated by Depth Anything 3~\cite{depthanything3}. 
Static Gaussians use the original 3DGS scale activation~\cite{3DGS}. We disable camera pose optimization. All other settings follow MoSca~\cite{MoSca}. All experiments are conducted on an NVIDIA RTX A6000 GPU.

\subsection{Baselines}
\label{subsec:baselines}
We benchmark our method against various 3DGS-based baselines: D3DGS~\cite{D3DGS}, FTGS~\cite{FreeTimeGS}, and MoSca~\cite{MoSca}.

To isolate the contribution of our multi-view consistent 3D track initialization, we construct naive multi-view extensions of MoSca, a state-of-the-art method for monocular dynamic scene reconstruction. These extensions follow the approach described in~\cref{subsec:naive} and use the same tracker~\cite{CoTracker3}, query density, regularization losses, and optimization settings as our method, differing only in 3D track initialization. Additionally, we construct MoSca-M, which uses MVTracker~\cite{rajic2025mvtracker} for multi-view 3D point tracking.

To validate the necessity of our initialization strategy, we compare against D3DGS~\cite{D3DGS}, a prior-free method for monocular dynamic-camera 4D reconstruction. D3DGS warps canonical Gaussians over time with a deformation MLP and is optimized using only photometric losses.

Since our sparse dynamic-camera setup has camera counts comparable to dense fixed-camera settings, we also compare against FTGS~\cite{FreeTimeGS}, a state-of-the-art method for dense fixed-camera 4D reconstruction that models complex motion by varying Gaussian positions and opacities.

For fair comparison, we prepare D3DGS* and FTGS* variants that use the same static/dynamic Gaussian separation as our method, with both variants using the original 3DGS~\cite{3DGS} for static Gaussians.

\subsection{Metrics}
\label{subsec:metrics}
We evaluate reconstruction quality via novel-view synthesis. Specifically, we train with three dynamic cameras and render the held-out fixed-camera view. We then compare rendered images with ground truth using PSNR, SSIM, and LPIPS~\cite{LPIPS}. We report each metric on the full image (Full) and on dynamic regions (Dynamic). Dynamic regions are defined as the union of the dynamic object mask from~\cref{sec:dataset} and an optical-flow-magnitude mask obtained by thresholding flow magnitude between adjacent frames.

\paragraph{Co-visibility mask} To evaluate only regions co-visible in the training and evaluation views, we apply a co-visibility mask to both rendered and ground-truth images before metric computation. Unlike~\cite{DyCheck}, which estimates co-visibility using forward-backward optical-flow consistency, we compute co-visibility by back-projecting points from training views into the evaluation view and checking visibility. This is important in our setting because the evaluation camera is farther from the training cameras, where flow-based checks tend to underestimate co-visible regions. Although this method can produce sparse masks in some cases, it more accurately identifies truly co-visible regions and therefore yields a more meaningful evaluation. Metric computation with masks otherwise follows~\cite{DyCheck}. We additionally apply several morphological opening and closing operations to obtain denser co-visibility masks. Following~\cite{HiMoR}, we compute metrics under the assumption that dynamic regions are always visible.

\paragraph{Cameraman exclusion} While cameramans are rarely seen in the training views, they are visible in the evaluation view. To assess the reconstruction quality of the target dynamic objects without bias from the cameraman, we exclude cameraman regions from metric computation. We use the dynamic object mask from~\cref{sec:dataset} to exclude cameraman regions.

\section{Experiments and Results}

\begin{figure*}
  \includegraphics[width=\linewidth]{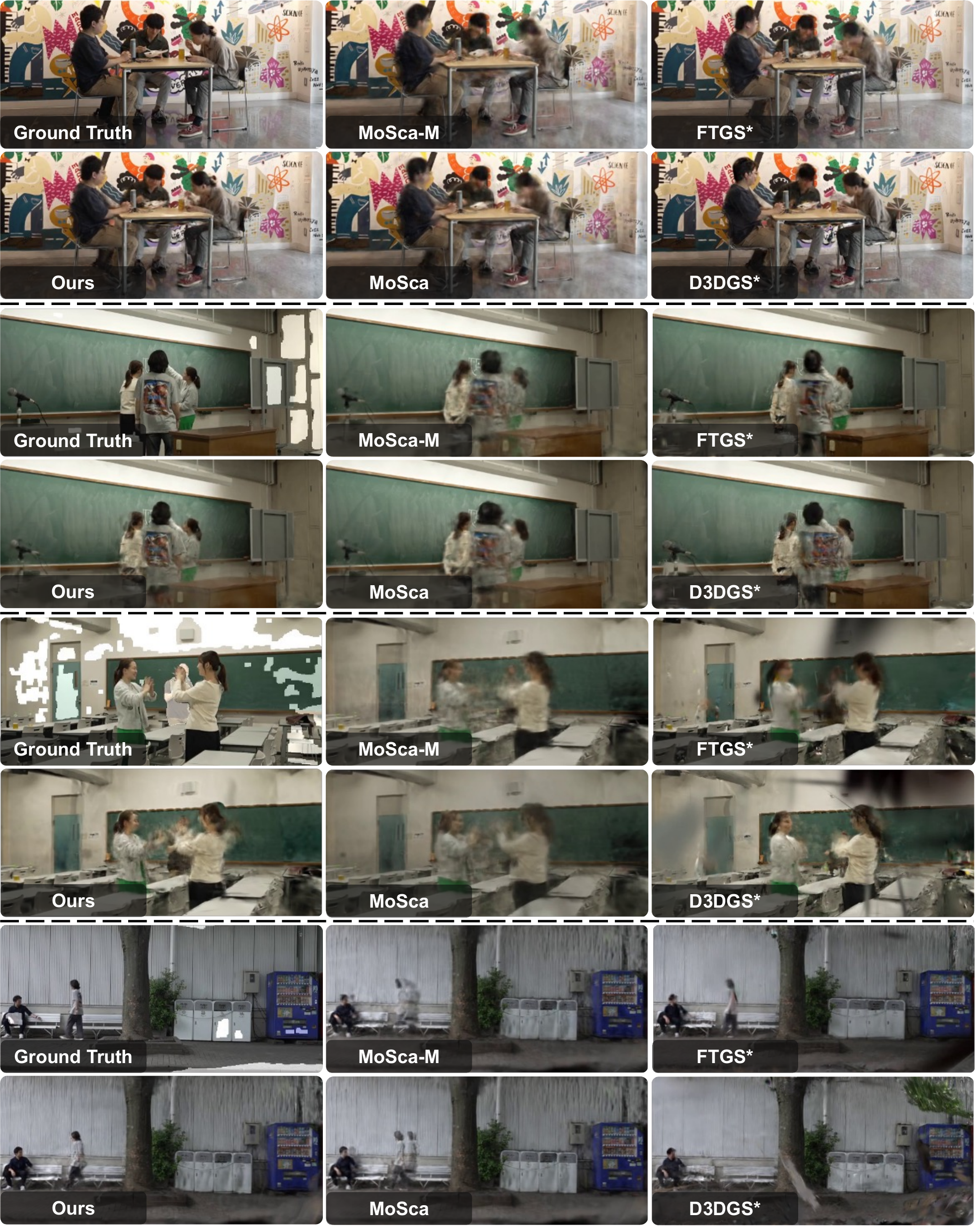}
  \caption{\textbf{Qualitative Results on \nickname{}}. Excluded regions from quantitative evaluation are masked in white.}
  \label{fig:qalitative}
\end{figure*}

\subsection{Benchmarking on \nickname{}}
\label{subsec:benchmarking}
All experiments are conducted at 30 FPS and at half FHD resolution. Quantitative and qualitative results on the \nickname{} benchmark are shown in Table~\ref{tab:quantitative}, \cref{fig:teaser}, and \cref{fig:qalitative}, respectively. Additional visualizations are provided in the supplementary videos. D3DGS is not included since it runs out of memory due to the significant increase in the number of Gaussians.

Compared to naive MoSca baselines, our method improves almost all metrics across scenes, with only a few SSIM exceptions. Gains are largest in difficult cases such as \textit{Lunch} (independent camera motion with large baselines) and \textit{Walking}/\textit{Bench} (large global motion of both cameras and subjects). We attribute this to our multi-view-consistent 3D track initialization: intra-camera tracking propagates correspondences over time and aggregates evidence from all frames, even when per-timestep multi-view cues are sparse. Although MVTracker is robust to noisy depth, it is trained mainly on fixed-camera datasets, so it generalizes poorly when both cameras and subjects move substantially. This is consistent with our quantitative results, where MoSca-M yields worse LPIPS than MoSca under dynamic settings.

FTGS* is competitive when baseline dynamics are small and initial correspondences are stable. Its flexible Gaussian deformation achieves strong pixel-level alignment, reflected by relatively high PSNR/SSIM in scenes such as \textit{Blackboard} and \textit{Bench}. However, because FTGS* does not explicitly leverage long-range temporal information, it struggles with fast motion and produces blurrier, less temporally consistent reconstructions (\cref{fig:qalitative}), along with consistently higher LPIPS than ours. By integrating visual information over all timesteps, our method produces sharper and more coherent results, especially under large baseline changes (e.g., \textit{Lunch}, \textit{Play}) and challenging camera trajectories (e.g., backward motion in \textit{Walking}), where it achieves higher PSNR and markedly lower LPIPS.

D3DGS* is robust to local motion but misses fine-grained dynamics. In \textit{Lunch}, our reconstruction better preserves the right-hand details of the central and rightmost subjects. In \textit{Play}, D3DGS* reconstructs torso appearance reasonably well but suppresses motion, resulting in near-static behavior. More importantly, D3DGS* significantly degrades reconstruction quality when subjects undergo large global motion. These results suggest that, in sparse dynamic-camera settings, having many viewpoints across time is still insufficient when warp-based optimization relies only on photometric loss and a simple motion field (e.g., an MLP). A flexible motion field (e.g., MoSca) and a strong initialization prior are both critical, and our framework with spatiotemporally consistent 3D-track initialization provides this prior effectively.

\subsection{Ablation Study}
\label{subsec:ablation}
All quantitative results in this section are averaged across all scenes.

\paragraph{3D Track Initialization} We ablate our spatiotemporal 3D track initialization: epipolar constrain check (Epi) and triangulation (Tri). Results are shown in Table~\ref{tab:ablation_epi_tri}. The table shows that the proposed epipolar constraint check and triangulation-based depth estimation are effective.

\paragraph{Depth Loss} We further ablate the depth loss design, as shown in Table~\ref{tab:ablation_depth_loss}. The results show that the ordinal depth loss is more effective than the normalized depth loss, and both are better than not using depth loss.

\paragraph{Batch Sampling} We further ablate the batch sampling strategy. Table~\ref{tab:ablation_batch_sampling} shows that sampling from different views and different timesteps yields the best overall performance. We hypothesize that this setting generalizes best during optimization because it exposes the model to broader spatiotemporal variation.

\paragraph{Camera Setup} We validate the effectiveness of the sparse dynamic camera setup by comparing it with a monocular 
dynamic-camera setup, where training uses the single camera that has the largest overlap with the evaluation camera among the three dynamic cameras. As shown in \cref{fig:teaser}, the monocular setup suffers from depth ambiguity
whereas the sparse dynamic camera setup alleviates these issues by providing additional viewpoints. Notably, a direct multi-view extension of MoSca performs qualitatively worse than the monocular baseline. In contrast, our method with spatiotemporally consistent 3D track initialization achieves substantially better results than both baselines.

\begin{table}[t]
  \centering
  \caption{Ablation on 3D track initialization. Epi denotes epipolar constraint check, and Tri denotes triangulation. \cmark/\xmark indicate whether each process is enabled.}
  \label{tab:ablation_epi_tri}
  \resizebox{\linewidth}{!}{%
  \begin{tabular}{cccccccc}
    \toprule
    & & \multicolumn{3}{c}{Full} & \multicolumn{3}{c}{Dynamic} \\
    Epi & Tri & PSNR$\uparrow$ & SSIM$\uparrow$ & LPIPS$\downarrow$ & PSNR$\uparrow$ & SSIM$\uparrow$ & LPIPS$\downarrow$ \\
    \midrule
    \xmark & \xmark & 18.46 & 0.773 & 0.297 & 15.40 & \second{0.984} & 0.246 \\
    \cmark & \xmark & 18.42 & 0.772 & 0.299 & 15.51 & \second{0.984} & 0.252 \\
    \xmark & \cmark & \second{18.71} & \second{0.777} & \second{0.292} & \second{17.65} & \best{0.987} & \second{0.184} \\
    \cmark & \cmark & \best{19.06} & \best{0.783} & \best{0.281} & \best{18.05} & \best{0.987} & \best{0.167} \\
    \bottomrule
  \end{tabular}
  }
\end{table}

\begin{table}[t]
  \centering
  \caption{Ablation on depth loss.}
  \label{tab:ablation_depth_loss}
  \resizebox{\linewidth}{!}{%
    \begin{tabular}{lcccccc}
      \toprule
      & \multicolumn{3}{c}{Full} & \multicolumn{3}{c}{Dynamic} \\
       & PSNR$\uparrow$ & SSIM$\uparrow$ & LPIPS$\downarrow$ & PSNR$\uparrow$ & SSIM$\uparrow$ & LPIPS$\downarrow$ \\
      \midrule
      W/o Depth Loss
        & 18.95 & 0.766 & \second{0.285}
        & 18.25 & \best{0.990} & 0.148 \\
      Normalized Depth Loss
        & \second{18.97} & \second{0.767} & 0.288
        & \second{18.27} & \best{0.990} & \second{0.147} \\
      Ordinal Depth Loss
        & \best{19.02} & \best{0.768} & \best{0.282}
        & \best{18.36} & \best{0.990} & \best{0.144} \\
      \bottomrule
    \end{tabular}
  }
\end{table}

\begin{table}[t]
  \centering
  \caption{Ablation of batch sampling strategies. S/D denote sampling from the same/different view and the same/different timestep, respectively.}
  \label{tab:ablation_batch_sampling}
  \resizebox{\linewidth}{!}{%
  \begin{tabular}{cccccccc}
    \toprule
    & & \multicolumn{3}{c}{Full} & \multicolumn{3}{c}{Dynamic} \\
    View & Time & PSNR$\uparrow$ & SSIM$\uparrow$ & LPIPS$\downarrow$ & PSNR$\uparrow$ & SSIM$\uparrow$ & LPIPS$\downarrow$ \\
    \midrule
    S & D & \second{18.92} & \second{0.767} & 0.290
        & 18.31 & \best{0.990} & \second{0.146} \\
    D & S & 18.86 & \second{0.767} & \second{0.286}
        & \second{18.33} & \best{0.990} & 0.148 \\
    D & D & \best{19.02} & \best{0.768} & \best{0.282}
        & \best{18.36} & \best{0.990} & \best{0.144} \\
    \bottomrule
  \end{tabular}
  }
\end{table}

\section{Conclusion}
\label{sec:conclusion}
This paper introduces 4D reconstruction from sparse dynamic cameras, a practical yet underexplored setup where a small number of independently moving cameras capture the same dynamic scene. To address the spatiotemporal inconsistencies that limit naive extensions of existing methods, we proposed a unified framework with three key components: multi-view consistent 3D track initialization, noise-robust depth-order regularization, and spatiotemporally diverse batch sampling.
To support rigorous evaluation in this setting, we presented \nickname{}, a real-world benchmark with five sequences across four diverse environments, captured by three dynamic cameras and one fixed evaluation camera. Comprehensive experiments showed that our method consistently improves reconstruction quality, especially in dynamic regions and challenging motion scenarios, compared with representative baselines. 

\paragraph{Acknowledgements} We would like to thank Yuki Asukabe, Ryosuke Yokoya, Kanae Hirao, Kohei Kawanishi, Tomoyuki Minani, Tomoaki Satsuka, and Hiroshi Arai for their assistance with dataset collection. 

{
    \small
    \bibliographystyle{ieeenat_fullname}
    \bibliography{main}
}

\clearpage
\setcounter{page}{1}
\maketitlesupplementary

\section{Details of \nickname{}}
\label{sec:dataset_details}

\subsection{Data Acquisition}
\label{subsec:data_acquisition}
\nickname{} is captured using three independently moving cameras that follow the subjects during recording, mimicking realistic handheld or operator-driven capture scenarios. We use two Sony FX3 cameras~\cite{FX3} and two Sony $\alpha$7S III cameras~\cite{A7S3}, where three cameras act as dynamic cameras and one $\alpha$7S III serves as a evaluation camera for benchmark. To ensure accurate camera intrinsics, we disable in-body image stabilization, which otherwise alters the effective optical center. Temporal synchronization across all cameras is achieved using UltraSync devices~\cite{UltraSync}, which transmit a shared timecode via Bluetooth and embed identical timestamps into each recording.

\subsection{Dataset Characteristics}
\label{subsec:characteristics}
\nickname{} features five sequences of realistic and challenging everyday activities across four diverse environments (classroom, lounge, bench, and road), producing a broad range of camera trajectories, viewpoints, and scene layouts. Table~\ref{tab:dataset_comparison} summarizes the key characteristics of \nickname{} and major existing dynamic scene video datasets. Similar to~\cite{DyCheck}, each sequence is relatively long (from 8 to 15 seconds), enabling long-term non-rigid motion and sustained interactions rather than short isolated actions~\cite{Neural3DV, NVIDIA}. The scenes cover motion from subtle behaviors (e.g., eating lunch or discussing at a blackboard) to dynamic actions (e.g., walking), and include 11 subjects interacting with each other and surrounding objects, yielding complex multi-object dynamics and frequent severe occlusions that highlight the importance of multi-view consistency.

\subsection{Annotaion details}
\label{subsec:annotation}
\paragraph{Camera poses} Accurate camera pose estimation is a critical prerequisite for high-quality 4D reconstruction. To allow full reproducibility, we use the standard Structure-from-Motion (SfM) pipeline, e.g., COLMAP~\cite{COLMAP}. The problem under the sparse dynamic camera setup is the large number of images. Specifically, the total number of images whose poses must be estimated scales as $N_{\text{cam}} \times N_{\text{time}}$, where $N_{\text{cam}}$ is the number of cameras and $N_{\text{time}}$ is the number of time steps. In contrast, dense or sparse fixed-camera setups require estimating only $N_{\text{cam}}$ poses, while monocular dynamic setups require estimating only $N_{\text{time}}$ poses. In our dataset, with three cameras recording at 60~FPS for 23~seconds, each sequence contains approximately 4{,}000 images. Estimating camera poses for such a large number of images, with exhaustive image matching quickly becomes computationally prohibitive and often leads to CPU memory exhaustion.

Hence, we exploit the known temporal ordering of the images and adopt a temporal locality-aware matching strategy inspired by~\cite{hierarchicalgaussians24}. Specifically, instead of exhaustively matching all image pairs, we restrict feature matching to temporally nearby frames across cameras. For each image captured at time step $i$, we only match it to images at time steps $i + 2^k$, where $k \in [0, 10]$, effectively covering both short- and long-range temporal correspondences while keeping the number of candidate pairs tractable. This temporally structured matching strategy substantially reduces computational cost while preserving sufficient multi-view geometric constraints for robust camera pose estimation. Obtained camera poses are visualized in Figure~\cref{fig:pose}.

\paragraph{Dynamic object masks} In addition to camera poses, we provide semi-automatically extracted dynamic object masks for all sequences. We generate these masks using SAM2~\cite{SAM2} by manually specifying prompt points for each dynamic object in a reference frame and automatically propagating the masks temporally across the sequence. This process is performed independently for each camera.

The resulting dynamic masks serve two purposes. First, they are used to exclude dynamic regions during camera pose estimation, improving the robustness of structure-from-motion. Second, they enable explicit static/dynamic separation during 4D reconstruction, which is essential for accurately modeling dynamic scenes.

\begin{table}[t]
  \centering
  \resizebox{\linewidth}{!}{%
    \begin{tabular}{lcccc}
      \toprule
      Dataset & \makecell[c]{Multiple\\Dynamic?} & \makecell[c]{Independently\\Moving?} & \makecell[c]{Common\\Targets?} & Real? \\
      \midrule
      N3DV~\cite{Neural3DV} & \xmark & - & \cmark & \cmark \\
      DyCheck~\cite{DyCheck} & \xmark & - & \cmark & \cmark \\
      nuScenes~\cite{nuScenes} & \cmark & \xmark & \xmark & \cmark \\
      ImVid~\cite{ImViD} & \cmark & \xmark & \cmark & \cmark \\
      Ego4D~\cite{Ego4D} & \cmark & \cmark & \xmark & \cmark \\
      Multi-Cam Video~\cite{ReCamMaster} & \cmark & \cmark & \cmark & \xmark \\
      LetCamsGo (ours) & \cmark & \cmark & \cmark & \cmark \\
      \bottomrule
    \end{tabular}%
  }
  \caption{\textbf{Comparison with major publicly available dynamic scene video datasets.} Most datasets feature either multiple fixed cameras or a single dynamic camera. When multiple dynamic cameras are present, they are typically rig-mounted, which limits camera motion independence; moreover, some datasets lack common targets across cameras or are not captured in real-world settings. LetCamsGo uniquely provides multiple independently moving cameras that share common targets in real-world scenarios.}
  \label{tab:dataset_comparison}
\end{table}


\end{document}